\documentclass{article} % For LaTeX2e
\usepackage{iclr2023_conference,times}

\usepackage{amsmath,amsfonts,bm}

\def\eqref#1{equation~\ref{#1}}
\def\1{\bm{1}}

\DeclareMathAlphabet{\mathsfit}{\encodingdefault}{\sfdefault}{m}{sl}
\SetMathAlphabet{\mathsfit}{bold}{\encodingdefault}{\sfdefault}{bx}{n}

\iffalse % don't show comments
    \newcommand\todo[1]{}

    \newcommand{\dumitru}[1]{}
    \newcommand{\mbz}[1]{}
    \newcommand{\pikinder}[1]{}
    \newcommand{\hmoraldo}[1]{}
    \newcommand{\msaffar}[1]{}
    \newcommand{\rubville}[1]{}
    \newcommand{\zhanghan}[1]{}
\else % show comments
    \newcommand{\todo}[1]{{\textcolor{red}{[[TODO: {#1}]]}}}

    \newcommand{\dumitru}[1]{\textcolor{magenta}{[dumitru: {#1}]}}
    \newcommand{\mbz}[1]{\textcolor{green}{[mbz: {#1}]}}
    \newcommand{\pikinder}[1]{\textcolor{blue}{[pikinder: {#1}]}}
    \newcommand{\hmoraldo}[1]{\textcolor{orange}{[hmoraldo: {#1}]}}
    \newcommand{\msaffar}[1]{\textcolor{gray}{[msaffar: {#1}]}}
    \newcommand{\rubville}[1]{\textcolor{purple}{[rubville: {#1}]}}
    \newcommand{\zhanghan}[1]{\textcolor{navy}{[zhanghan: {#1}]}}
\fi

\usepackage{hyperref}
\usepackage{url}
\usepackage{xspace}
\usepackage{ dsfont }
\usepackage{afterpage}
\usepackage{enumitem}

\usepackage{algorithm}
\usepackage{algorithmic}

\usepackage[export]{adjustbox}
\usepackage{makecell}
\usepackage{siunitx}
\usepackage{multirow}
\usepackage{extdash}

\usepackage{subcaption}
\usepackage{graphicx}
\usepackage{wrapfig}
\usepackage[capitalize]{cleveref}
\usepackage[symbol]{footmisc}

\newcommand{\model}{Phenaki\xspace}
\newcommand{\phenaki}{\model}
\newcommand{\website}{\href{https://phenaki.github.io/}{phenaki.github.io}}

\newcommand{\vivit}{ViViT\xspace}
\newcommand{\cvivit}{C-\vivit\xspace}
\newcommand{\story}{story\xspace}

\newcommand{\bx}{\mathbf{x}}
\newcommand{\bz}{\mathbf{z}}
\newcommand{\bp}{\mathbf{p}}

\newcommand{\be}{\mathbf{e}}

\title{\model: Variable length video generation\\from open domain textual descriptions} 

\iclrfinalcopy

\author{
Ruben Villegas\footnote[3]{}\\
Google Brain \\
\small{\texttt{rubville@google.com}} \\
\And
Mohammad Babaeizadeh\footnote[3]{}\\
Google Brain \\
\small{\texttt{mbz@google.com}} \\
\And
Pieter-Jan Kindermans\footnote[3]{}\\
Google Brain \\
\small{\texttt{pikinder@google.com}} \\
\AND
Hernan Moraldo\\
Google Brain \\
\small{\texttt{hmoraldo@google.com}} \\
\And
Han Zhang\\
Google Brain \\
\small{\texttt{zhanghan@google.com}} \\
\And
Mohammad Taghi Saffar\\
Google Brain \\
\small{\texttt{msaffar@google.com}} \\
\AND
Santiago Castro\footnote[1]{}\\
University of Michigan \\
\small{\texttt{sacastro@umich.edu}} \\
\And
Julius Kunze\footnote[1]{}\\
University College London \\
\small{\texttt{kjulius@google.com}} \\
\And
Dumitru Erhan\\
Google Brain \\
\small{\texttt{dumitru@google.com}} \\
\AND
}

\begin{document}

\maketitle
\footnotetext[3]{Equal contribution. * Intern at Google Brain while working on this project.}

\begin{abstract}
We present \model, a model capable of realistic video synthesis, given a sequence of textual prompts. Generating videos from text is particularly challenging due to the computational cost, limited quantities of high quality text-video data and variable length of videos. To address these issues, we introduce a new model for learning video representation which compresses the video to a small representation of discrete tokens. This tokenizer uses causal attention in time, which allows it to work with variable-length videos. 
To generate video tokens from text we are using a bidirectional masked transformer conditioned on pre-computed text tokens. The generated video tokens are subsequently de-tokenized to create the actual video. To address data issues, we demonstrate how joint training on a large corpus of image-text pairs as well as a smaller number of video-text examples can result in generalization beyond what is available in the video datasets. Compared to the previous video generation methods, \model can generate arbitrary long videos conditioned on a sequence of prompts (i.e. time variable text or \textit{a \story}) in open domain. To the best of our knowledge, this is the first time a paper studies generating videos from time variable prompts. In addition, compared to the per-frame baselines, the proposed video encoder-decoder computes fewer tokens per video but results in better spatio-temporal consistency.

% \pikinder{Sep 20: Me thinking out loud.
% \begin{itemize}
%     \item First effective video approaches NUWA and CogVideo rely on a per frame encoding.
%     \item VQGAN sota for image, but artefacts in video -> CVIVIT
%     \item Video data is limited, therefore need to train on image and text -> CVIVIT
%     \item Video sequences are long -> Generation with transformers slow -> Maskgit, CVIVIT
%     \item Story mode is not in the abstract
% \end{itemize}
% }
\end{abstract}
\section{Introduction}
\afterpage{%
\begin{figure}[htb]
  \centering
  \vspace{-0.8cm}
  \includegraphics[width=0.9\linewidth]{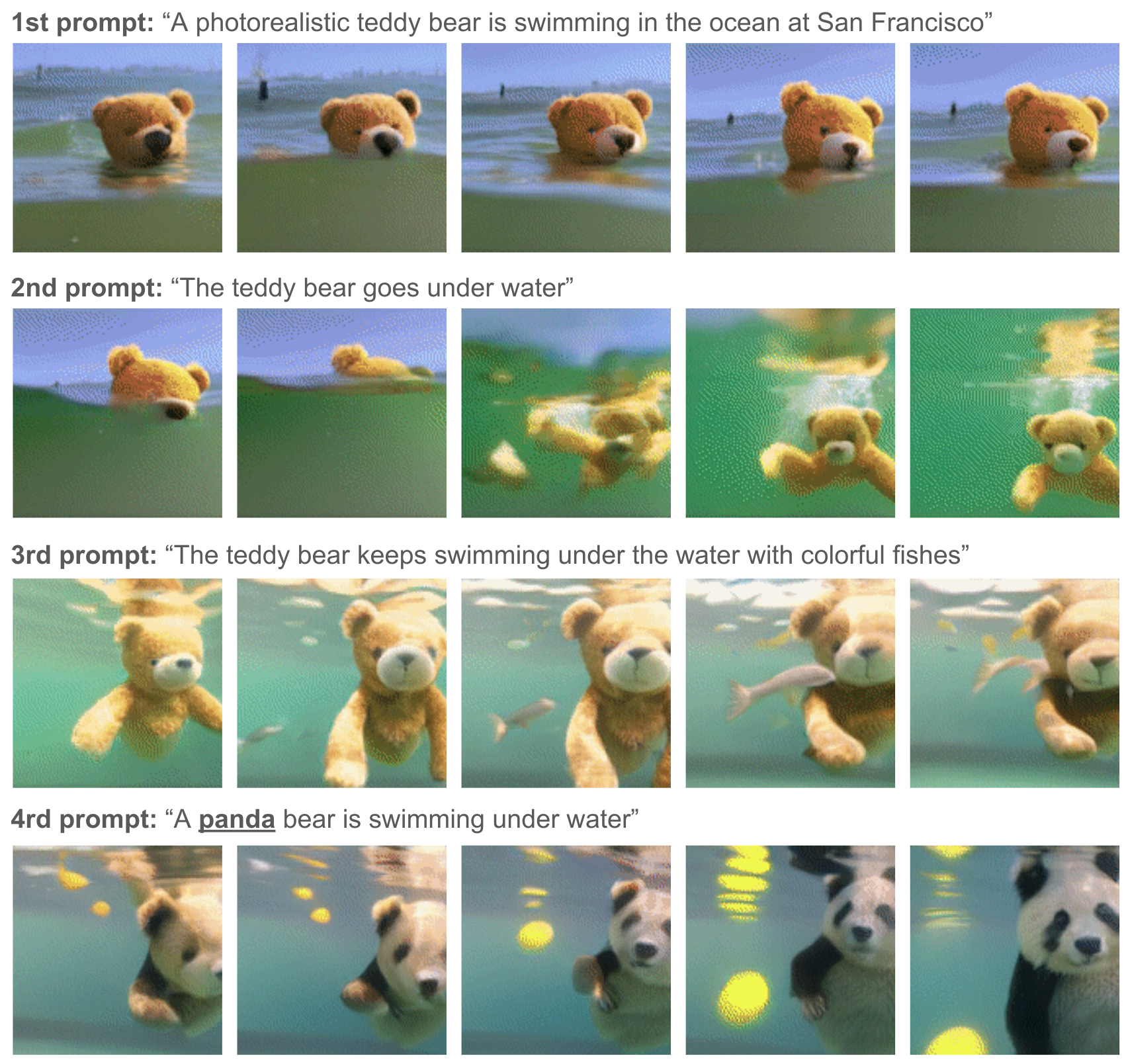}
  \caption{Time variable text (i.e. \story) conditional video generation. The entire figure is \textbf{one continuous video} generated auto-regressively. We start by generating the video conditioned on the first prompt and then after a couple of frames we change the prompt to the next one. Each row contains a selected number of frames (from left to right in order) while the model was conditioned on that particular prompt. The model manages to preserve the temporal coherence of the video while adapting to the new prompt, usually taking the shortest path for the adaption (notice the \textit{morphing} of the teddy bear to the panda). Please note that the generated video has complex visual features such as reflections, occlusions, interactions and scene transitions. Full video is available at \website.}
  \label{fig:story}
  \vspace{0.5cm}
\end{figure}%
}

It is now possible to generate realistic high resolution images given a description \cite{DALLE,ramesh2022hierarchical,nichol2021glide,saharia2022photorealistic,PARTI}, but generating high quality videos from text remains challenging. In essence, videos are just a sequence of images, but this does not mean that generating a long coherent video is easy.  In practice, it is a significantly harder task because there is much less high quality data available and the computational requirements are much more severe~\cite{clark2019adversarial}. For image generation, there are datasets with billions of image-text pairs (such as LAION-5B~\cite{schuhmann2021laion} and JFT4B~\cite{zhai2022scaling}) while the text-video datasets are substantially smaller~e.g. WebVid~\cite{bain2021frozen} with ${\sim}10$M videos, which is not enough given the higher complexity of open domain videos. As for computation, training current state-of-the-art image generation models is already pushing the state-of-the-art computational capabilities~\cite{PARTI}, leaving little to no room for generating videos, particularly videos of variable length.

To make the matters worse, one can argue that a single short text prompt is not sufficient to provide a complete description of a video (except for short clips), and instead, a generated video must be conditioned on a sequence of prompts, or \textit{a \story}, which narrates what happens over time. Ideally, a video generation model must be able to generate videos of arbitrary length, all the while having the capability of conditioning the generated frames at time $t$ on prompts at time $t$ that can vary over time. Such capability can clearly distinguish the \textit{video} from a ``moving image" and open up the way to real-world creative applications in art, design and content creation. To the best our knowledge, \story based conditional video generation has never been explored before and this is the first paper to take early steps towards that goal. A traditional deep learning approach of simply learning this task from data is not possible, since there is no story-based dataset to learn from. Instead, to achieve this we rely on a model that is designed specifically with this capability in mind.

% In this paper we introduce the \cvivit model with the following advantages:
In this paper, we introduce \phenaki, a text to video model trained on both text to video and text to image data that can:
\begin{itemize}[leftmargin=0.2in]
    \item[--] Generate temporally coherent and diverse videos conditioned on open domain prompts even when the prompt is a new composition of concepts (Fig.~\ref{fig:compos}). The videos can be long (minutes) even though the model is trained on 1.4 seconds videos (at 8 fps).
    \item[--] Generate videos conditioned on a \story (i.e. a sequence of prompts), e.g.~Fig.~\ref{fig:story} and Fig.~\ref{fig:story2}.
    % \item[--] it can also achieve competitive performance on traditional video generation tasks, despite not being trained on them explicitly.
\end{itemize}

To enable these capabilities, we could not rely on current video encoders, because they either can only decode fixed size videos or they encode frames independently. Hence, we introduce \cvivit, a novel encoder-decoder architecture that:
\begin{itemize}[leftmargin=0.2in]
\item[--] Exploits temporal redundancy in videos to improve reconstruction quality over a per frame model while compressing the number of video tokens by 40\% or more.
\item[--] Allows encoding and decoding of variable length videos given its causal structure.
\end{itemize}

%In this paper, we introduce \model, a model capable of generating videos from a \story i.e. a sequence of prompts. In order to address the computational requirements, we propose \cvivit, a causal variation of \vivit~\cite{vivit} with additional architectural design changes for video generation, which provides temporal-spatial compression while being auto-regressive in time. This enables the model to generate long videos while maintaining temporal coherence. To address data issues, we train the \model on a mix of images and videos. We demonstrate how this enables the model to \textit{transfer} motion from a smaller set of observed videos to new subjects that only exist in the image datasets. As such, \model can generalize to a wider domain compared to what text-video datasets provide. We also show for auto-regressive models, it is possible to generalize from single prompt description per video to a \story in zero-shot. This combination enables \model to generate videos that could span over multiple minutes, given a series of prompts that potentially describe a story. To the best of our knowledge, this is the first paper taking steps toward generating a long video based on time variable text prompts. 
\begin{figure}
  \centering
  \vspace{-0.5cm}
  \includegraphics[width=1\linewidth]{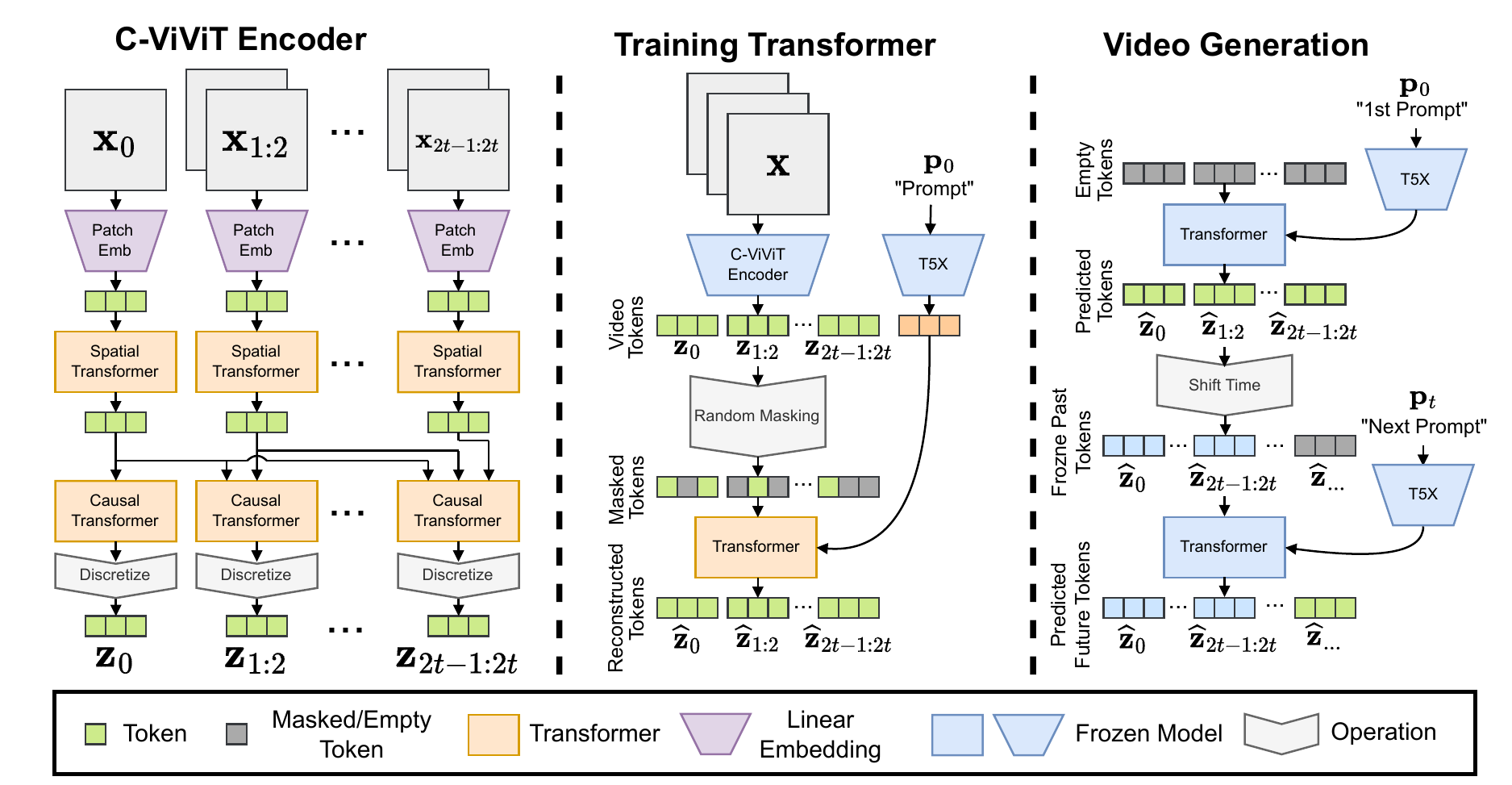}
  \vspace{-.1in}
  \caption{The architecture of \model. \textbf{Left:} \cvivit encoder architecture. The embeddings of images and video patches from raw frames $\bx$ are processed by a spatial and then a causal transformer (auto-regressive in time) to generate video tokens $\bz$. \textbf{Center:} MaskGiT is trained to reconstruct masked tokens $\bz$ predicted by a frozen \cvivit encoder and conditioned on T5X tokens of a given prompt $\bp_0$. \textbf{Right:} How \model can generate arbitrary long videos by freezing the \textit{past} token and generating the future tokens. The prompt can change over time to enable time-variable prompt (i.e. \story) conditional generation. The subscripts represent time (i.e. frame number).}
  \label{fig:models}
\end{figure}

\section{The \phenaki model}
\label{sec:method}
Inspired by the previous work in auto-regressive text to image~\cite{DALLE,PARTI,saharia2022photorealistic} and text to video~\cite{nuwa,wu2022nuwa,cogvideo}, \model is designed with two main components (see Figure~\ref{fig:models}): an encoder-decoder model which compresses videos to discrete embeddings (i.e. tokens) and a transformer model to \textit{translate} text embeddings to video tokens. To get the text embeddings, \model uses a pre-trained language model, T5X~\cite{roberts2022scaling}. We will discuss each one of these components in the following subsections.
 
% \subsection{Text embeddings}
% \rubville{Depending on results, talk about either learned text encoder or pre-computed t5x embeddings}
% \begin{itemize}
%     \item Reference to this model?
%     \item Sequence length is 64 tokens
%     \item embedding size is 4096
%     \item motivate using a pre-trained text tokenizer that it saves us memory since we can pre-compute embeddings and we do not need to load it into memory. This frees up space for compute. Imagen also had good results with it.
% \end{itemize}
%and trained key components that that work together to generate video from a text prompt. Additionally there is also a pre-trained text tokenizer based on T5X.
%The first component is the video (de-)tokenizer. It is able to take a video and return a compressed tokenized representation of this video. 
%text to video. The first model is an encoder-decoder transformer that quantizes video into sp89atio-temporal tokens. The second model maps textual descriptions into the spatio-temporal tokens computed using the encoder/decoder model. The tokens generated by the second model are then passed to the decoder in the first model to generate video. Next, we describe the details of each of our models, training strategies to learn from image and video datasets, super-resolution models, and other contributions that enable high-performing open domain text-to-video generation.

\subsection{Encoder-decoder video model: \cvivit}
% \rubville{Mention difference between the original vivit architecture and our non-causal \vivit architecture. Make sure reviewers don't think it's exactly the same.}
One of the primary challenges for generating video from text, is to get a compressed representation of videos. Previous work on text to video either use per-frame image encoders~\cite{cogvideo,nuwa,harp} such as VQ-GAN~\cite{vqgan} or fixed length video encoders~\cite{godiva} such as VideoVQVAE~\cite{videovqvae}. The former allows for generating videos of arbitrary length, however in practice, the videos have to be short because the encoder does not compress the videos in time and the tokens are highly redundant in consecutive frames. The latter is more efficient in the number of tokens but it does not allow to generate variable length videos. In \model, our goal is to generate videos of variable length while keeping the number of video tokens to a minimum so they can be modeled with a transformer within current computational limitations. To do so, we introduce \cvivit, a causal variation of \vivit~\cite{vivit} with additional architectural changes for video generation, which can compress the videos in temporal and spatial dimensions, while staying auto-regressive in time, This capability allows for generating videos of arbitrary length auto-regressively.
% given the reduuse a per frame image encoder to get video, who relied on a per frame model, as is typically used in text to image research. \pikinder{Can someone get us an illustration of this this}\rubville{It is difficult to notice the differences in a pdf. The FVD metric should be enough to show this}. However this approach has 2 key issues. First it does not use the redundancy between frames to encode information effectively. Second, when frames are decoded independently it can introduce temporal artefacts. We will discuss our basic ViVit encoder-decoder model first and discuss its limitations for autoregressive extension of video, which prevents it from being used in story mode. Then we will introduce C-ViVit that re-enables autogressive video generation.

\paragraph{Encoder architecture:}
As illustrated in Figure \ref{fig:models}, we start with a video sequence of $t_x + 1$ frames with a resolution of $w_x\times h_x$ and $c_x$ channels: $\bx \in \mathds{R}^{(t_x + 1) \times h_x \times w_x \times c_x}$. This sequence will be compressed into a token representation of size $(t_z + 1)\times w_z \times h_z$ where the first $w_z\times h_z$ tokens represent the first frame independently from the rest of the video, and the remaining tokens represent spatio-temporal video tokens that auto-regressively depend on previous frames. 
To do so, we extract non-overlapping image patches of size $w_p \times h_p \times c_p$ from the first frame and video patches of size $t_p\times w_p \times h_p \times c_p$ from the rest of the video.
We typically use all channels at once such that the number of patches equals the number of video tokens $t_z=\frac{t_x}{t_p}$, $w_z=\frac{w_x}{w_p}$ and  $h_z=\frac{h_x}{h_p}$.
Each of these patches is flattened and linearly projected into a $d_z$ dimensional space.
We combine the spatial dimensions to have a tensor of shape $(t_z + 1)\times w_z*h_z \times d_z$ where the spatial and temporal dimensions are separated.
Then multiple transformer layers are applied along the spatial dimensions with all-to-all attention.
This is followed by multiple transformer layers over the temporal dimension with causal attention such that each spatial token only observes spatial tokens from previous frames in an auto-regressive manner.
The effect of this is that the first frame can be completely independently encoded.
This opens up the possibility of text to image training to be embedded naturally into our video model.
The second advantage is that we can condition the video generation process on a number of starting frames. 
The resulting patch embeddings $\bz$ of shape $t_z\times w_z \times h_z \times d_z$ are then tokenized into learned codewords $c_z$ by vector quantization.
The codebook learning will be discussed later together with the losses.
% \pikinder{Expand this to explain the genius and the novelty of this}

\paragraph{Decoder architecture:}
The \cvivit decoder is simply an upside down version of the encoder. First tokens are transformed into embeddings. This is followed by the temporal transformer, then the spatial transformer.
After the output of the spatial transformer, we apply a single linear projection without activation to map the tokens back to pixel space.

\paragraph{Quantization and Losses:}
To learn a discrete latent space, we quantize our encoder outputs into the entries of a learned codebook via the vector quantization (VQ) objective in VQVAEs \citep{vqvae},
\begin{equation}
    L_{\textit{VQ}} = \| \text{sg}(\bz) - \be \|^2_2 + \beta \| \bz - sg(\be) \|^2_2,
\end{equation}
where $\text{sg}(x) \equiv x$, and $\frac{\text{d}}{\text{d}x} \text{sg}(x) \equiv 0$ is the stop-gradient operator, $\beta$ is the commitment loss weight, and $\be$ is a codebook vector from codebook $\mathbf{E}$. The index to the codebook vector closest to $\bz$ is found by $i = \text{argmin}_j \| \bz - \mathbf{E}_j \|^2_2$. 
In addition to the VQ objective, we adopt the factorized and $\ell_2$-normalized codes from ViT-VQGAN~\citep{vit_vqgan} to improve codebook usage and reconstruction quality.

To train our model, we use a combination of $L_2$ loss, image perceptual loss $L_{\textit{IP}}$~\citep{JohnsonPerceptual2016,ZhangPerceptual2018}, video perceptual loss $L_{\textit{VP}}$ by using the I3D network~\citep{i3d} as feature extractor, and adversarial loss $L_{\textit{Adv}}$ with StyleGAN architecture~\citep{stylegan}.
As training objective, we use the following
\begin{equation}
    L = L_{\textit{VQ}} + 0.1\times L_{\textit{Adv}} + 0.1\times L_{\textit{IP}} + 1.0\times L_{\textit{VP}} + 1.0\times L_2.
\end{equation}

\paragraph{Novelty over the \vivit architecture:}
While our proposed \cvivit architecture is inspired by the factorized encoder in \vivit~\cite{vivit}, we modify their architecture to enable self-supervised learning from unlabeled videos. We first remove the \texttt{[CLS]} tokens in the spatial and the temporal transformers. Next, we apply temporal transformer for all spatial tokens computed by the spatial encoder, in contrast to  single run of the temporal transformer over the \texttt{[CLS]} tokens in \vivit.
% We denote the modification of their factorized encoder as \vivit encoder.
Most importantly, the \vivit encoder requires a fixed length video input due to the all-to-all attention in time. Therefore, we apply causal attention instead such that our \cvivit encoder becomes auto-regressive and allows for a variable number of input frames which are necessary to learn from image datasets, and auto-regressively extrapolate video or single frames into the future.

\subsection{Text-to-video generation with bidirectional transformers}
In this stage, the text-to-video task can be formulated as a sequence-to-sequence problem to predict video tokens given the paired text embeddings. Most of recent methods \citep{DALLE, PARTI, nuwa, cogvideo} adopt a  transformer model for these sequence-to-sequence tasks. In their models, they use an auto-regressive transformer which predicts the image or video tokens sequentially given the encoded text features.
% The sampling time scales linearly with the output size and is very slow for high-resolution images and long video sequences.
As a result, the sampling time scales linearly with the sequence length, even when caching is used. This becomes impractical for long video sequence generation. 

\paragraph{Masked bidirectional transformer:}
% \paragraph{Advantages}
% \begin{itemize}
%     \item Inference time
%     \item Quality-time trade off
% \end{itemize}

In this work, we aim to reduce the sampling time by having a small and fixed sampling step disregarding different video sequence lengths. Inspired by previous work for image generation~\citep{MASKGIT}, we use a bidirectional transformer since it can predict different video tokens simultaneously. For training step $i$, we first sample a mask ratio $\gamma_{i}$ from 0 to 1 and randomly replace $\lceil \gamma_{i} \cdot N \rceil$ tokens with the special token \texttt{[MASK]}, where $N$ is the video sequence length. Then
we learn the model parameters by minimizing the cross entropy loss on those masked tokens given the encoded text embeddings and unmasked video tokens. During inference, we first label all of the video tokens as the special token \texttt{[MASK]}. Then, at each inference step, we predict all the masked (unknown) video tokens in parallel conditioned on the text embeddings and unmasked (predicted) video tokens. We keep a ratio $\beta_{i}$ of the predicted tokens at sampling step $i$ and the remaining tokens are re-masked and re-predicted in the next step. 

As discussed in MaskGIT~\citep{MASKGIT}, the masking schedule $\gamma_{i}$ and sampling schedule $\beta_{i}$ have a significant effect on the samples quality therefore we follow the same strategies. Compared to an auto-regressive transformer, the number of sampling steps is an order-of-magnitude smaller (typically we use values in the range of 12 to 48). Generally speaking, more sampling steps improves the quality.

% \subsection{Learning a single model from image and video datasets}
\paragraph{Losses and training strategies:}
% \rubville{Should we use a different notation for the token indices for maskgit training?}
Given a pre-trained \cvivit, videos are encoded into codebook ids $\textbf{a}$ of shape $(t_z + 1)\times w_z \times h_z$ which are flattened into a long vector using the raster ordering from \cite{vit_vqgan}. We then model the text-conditional video token distribution using \textit{Masked Visual Token Modeling} (MVTM)~\cite{MASKGIT}:
\begin{equation}
    L_{\text{mask}}= - \sum\nolimits_{\forall i\in[1,N],m_i=1} \log p(a_i | \textbf{a}_{\bar{M}}, \bp),
\end{equation}
where $\textbf{a}_{\bar{M}}$ represents the masked version of $\textbf{a}$, $m_i$ is a binary variable indicating whether $a_i$ is masked or not, $N$ is the number of video tokens, and $\bp$ is the text condition embedding.
In addition to the MVTM objective, we train using classifier-free guidance by dropping the text condition $10\%$ of the time during training~\cite{cfg,PARTI} .
Finally, we dynamically adjust the MVTM objective during training to allow the use of image and video datasets as a single large dataset.
We achieve this by only applying the masking ratio and objective on the first $w_z \times h_z$ tokens if only a single frame is given or over all video tokens if a full video is given. 
This mixed image and video dataset training strategy allows our models to learn concepts only present in image datasets, and transfer them to concepts present video datasets (e.g., the pencil drawing styled video of the panda in Figure.\ref{fig:compos}).

\paragraph{Inference and auto-regressive generation of long videos:}
At inference time, we sample videos tokens by the same iterative process used in \cite{MASKGIT} with classifier-free guidance scale $\lambda$ to control alignment between the generation and the text condition. Once the first video is generated, we can extrapolate additional frames auto-regressively by encoding the last $K$ generated frames in the last video using \cvivit, initializing MaskGIT with the tokens computed by our \cvivit encoder, and proceed to generate the remaining video tokens conditioned on a text input. During video extrapolation, the text condition can be the same or a different one which enables our model to dynamically create visual transitions between the previous and current text condition visual content, effective generating a visual story an described by the input text.

% \subsection{High fidelity video super resolution}
% In order to increase the fidelity of generated videos, we use video super resolution model BasicVSR~\cite{chan2021basicvsr} and video frame interpolation model CAIN~\cite{choi2020cain}. We train BasicVSR from scratch on a mix of ground truth and generated videos with default hyper-parameters at \url{shorturl.at/ruWY4}. For CAIN, we used the pre-trained checkpoint at \url{shorturl.at/DOZ06}. These models double the spatial and temporal resolution of videos generated by \model.
\section{Experiments}
\label{sec:exp}
To evaluate \model, we test it on the following tasks: 1)~text conditional video generation, 2)~text-image conditional video generation, 3)~time variable text conditional video generation (i.e.) \story mode, 4)~video quantization and 5)~image conditional video generation a.k.a. video prediction. To the best of our knowledge, 3)~time variable text conditional video generation has not been explored in prior work. Given the dynamic nature of videos, we highly encourage readers to visit \website\ to check the generated videos. The website also includes qualitative comparisons to a subset of the prompts from the CogVideo paper~\cite{cogvideo}. While the focus is on the text to video generation tasks, it is remarkable that \phenaki is still competitive on the more traditional video tasks despite not being developed explicitly for these tasks. We implemented \phenaki in JAX~\cite{jax2018github} using FLAX~\cite{flax2020github} library.
% \rubville{Not sure where exactly to include text+frame-conditioned generation but we have to include it}
% In this section, we evaluate our method in three key aspects: 1) video reconstruction quality of our encoder-decoder model, 2) frame-conditioned video generation performance without text conditioning, 3) text-conditioned and text+frame-conditioned video generation, and 4) visual story generation by dynamic text inputs. We encourage the readers to visit \website, and check out our video samples.
\subsection{Text conditional video generation} 
\label{sec:t2v}
Currently there is no established benchmark for evaluating text to video methods.
This makes comparing \phenaki to recent methods such as NUWA~\cite{nuwa}, CogVideo~\cite{cogvideo}, NUWA-Infinity~\cite{wu2022nuwa} and video diffusion models \cite{ho2022video} difficult. 

Unless specified otherwise, we train a 1.8B parameter \phenaki model on a corpus of ${\sim}15$M text-video pairs at 8 FPS mixed with ${\sim}50$M text-images plus ${\sim}400$M pairs of LAION-400M~\cite{schuhmann2021laion} (more details in Appendix~\ref{app:sec:text2vid}).  The model used in the visualisations in this paper was trained for 1 million steps at a batch size of 512, which took less than 5 days. In this setup 80\% of the training data came from the video dataset and each image dataset contributed 10\%.  

\begin{figure}[htb]
  \centering
  \includegraphics[width=1\linewidth]{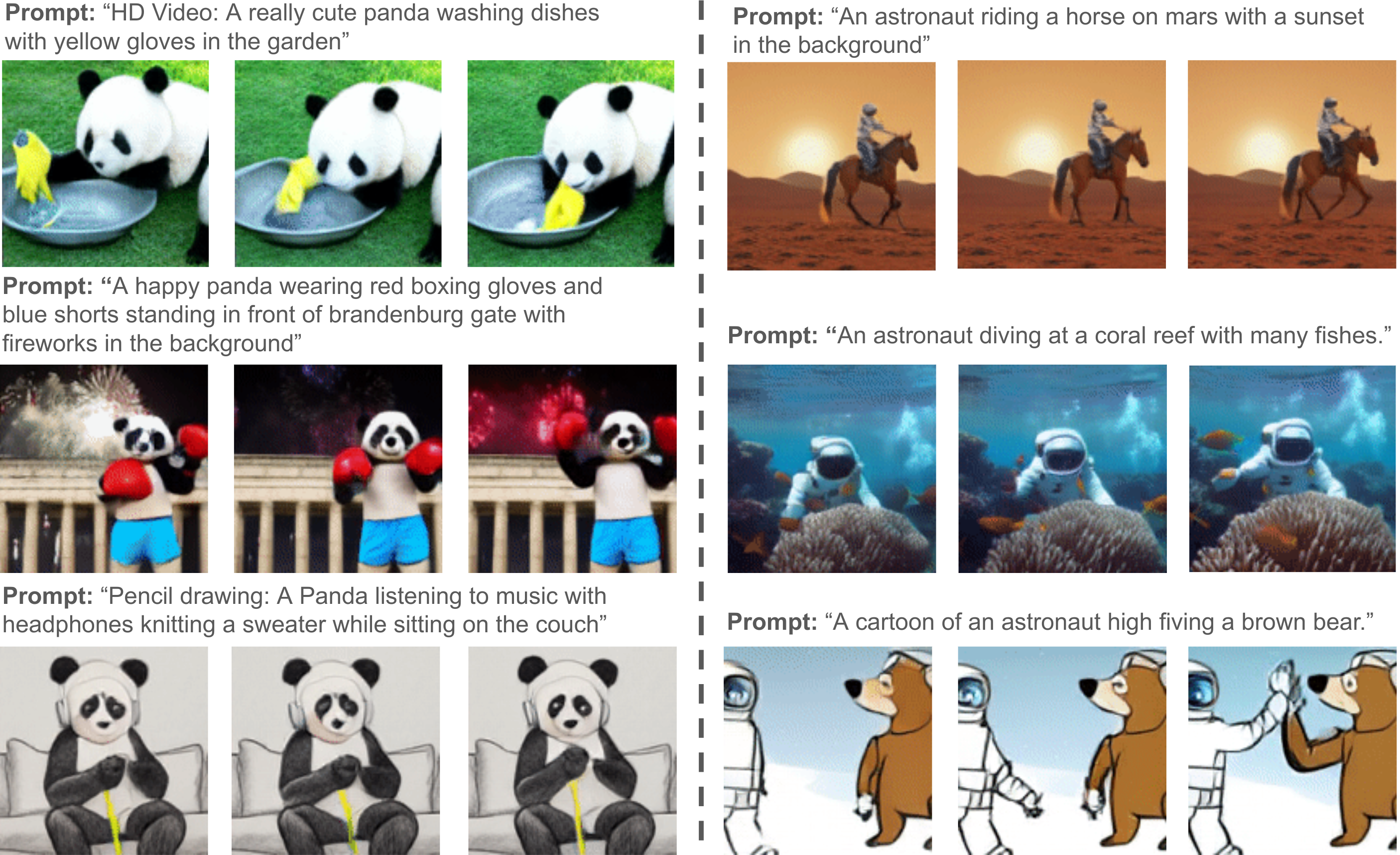}
  \caption{Text conditional video generation. Each row shows selected frames from a video generated given the prompt. The model is trained on a mix of images and videos. The video dataset does not include any \textit{stylized} videos such as pencil drawings, however, the image dataset does. The model can generalize from still images to videos. This figure also demonstrate the capability of the model in generating new unseen compositions. Full videos are available at \website.}
  \label{fig:compos}
\end{figure}

\paragraph{Qualitative evaluation:} Samples from this model can be seen in Figure~\ref{fig:compos} and additional samples are provided at \website. We observe that there is a high degree of control over both the actors and the background dynamics in the videos. The appearance of the actors and the video style can be adjusted by the text prompt as well (e.g. a regular video, a cartoon or a pencil drawing). 

On \website\ we provide examples from prompts that were provided in the CogVideo~\cite{cogvideo} demo. Since there are substantial differences between these methods it is hard to compare them on an equal footing. As an example, there are massive differences in scale: 9B parameters for CogVideo and 1.8B for our model. Additionally, the training data is different. Finally, we do not know how representative the prompts in the CogVideo demo are for the general performance of the CogVideo. 

\paragraph{Quantative comparison:} 
The NUWA~\cite{nuwa} paper provided a qualitative evaluation on Kinetics-400. Since the NUWA model is only $0.9$B parameters we also use a model of the same size. Our model was trained on 50\% video and 50\% image data in this experiment. The NUWA model fine-tuned on Kinetics but the \phenaki model is not: it is evaluated in a \emph{zero shot setting}.  The results in Table~\ref{table:t2v_comp} show that \phenaki achieves comparable generation quality, in a zero-shot setting, compared to previous text to video methods that were actually trained or finetuned on this dataset. 
% \rubville{The numbers here keep improving and closing in on the FID-img numbers :P. I need to specify the temperature, cfg scale, and maskgit iterations being used here}

\paragraph{On the importance of joint text-to-image and text-to-video training}
\begin{table}
    \small
    % \vspace{-0.5cm}
    \setlength{\tabcolsep}{2pt}
    \begin{minipage}{.3\linewidth}
        \caption{\small Text to video comparisons on Kinetics-400~\cite{Kinetics400}.}
        \label{table:t2v_comp}
        \centering
        \begin{tabular}{lrr}
        Method  & \begin{tabular}[x]{@{}c@{}}FID\\Image\end{tabular} $\downarrow$ & \begin{tabular}[x]{@{}c@{}}FID\\Video\end{tabular} $\downarrow$ \\
        \hline
        T2V \cite{vidgenfromtext}     & 82.13 & 14.65 \\
        SC \cite{tfgan}                 & 33.51 & 7.34 \\
        TFGAN \cite{tfgan}                   & 31.76 & 7.19 \\
        NUWA                      & 28.46 & 7.05 \\
        \hline
        Phenaki [0-Shot]            & 37.74 & 3.84 \\
        \end{tabular}
    \end{minipage}\hspace{1.5cm}%
    \begin{minipage}{.6\linewidth}
        \caption{\small Text to video and text to image results highlighting the importance of image datasets in video models. Text-to-image evaluation is done on ${\sim}40$K images of LAION-400M~\cite{schuhmann2021laion}.
        %\rubville{cfg=6, temp=4, iterations=24, over 4096 videos from lms}
        }
        \label{tab:video_image}
        \centering
        \begin{tabular}{c|lll|ll}
        Data Split & \multicolumn{3}{c|}{Text to Video} & \multicolumn{2}{c}{Text to Image} \\\hline\hline
        Vid\% / Img\% & CLIP $\uparrow$ & FID $\downarrow$ & FVD $\downarrow$ & CLIP $\uparrow$ & FID $\downarrow$ \\\hline
        100\% / 0\% & 0.298 & 19.2 & 168.9 & 0.240 & 53.9\\
        80\% / 20\% & 0.303 & 21.4 & 198.4 & 0.289 & 29.4\\
        50\% / 50\%& 0.302 & 21.4 & 239.7 & 0.287 & 30.5\\             
        \end{tabular}
    \end{minipage}
    
\end{table}
While there are some text-video datasets, text-image datasets dominate the internet in terms of quality and quantity \cite{videocc}. Consequently, there is simply not enough video data available to cover all the concepts present in text-image datasets. For example using only our video data, concepts such as pencil drawings or different painting styles cannot be learned. To be able to learn a model that can combine video dynamics with these additional concepts we have to combine training on image and video data. In Table~\ref{tab:video_image}, we evaluate the performance of using different ratios of video and images. We start with data splits of only video, and vary the ratio of image and video datasets up to using $50\%$ image and $50\%$ video datasets. In our results, we find that there is a trade-off in performance between models trained with only video video (i.e., significantly better FVD), and models trained with more image data (i.e., better text-video and text-image alignment, and significantly better FID in image datasets). On \website\ we show samples from different models side by side where this trade-off between control over the content and the quality of the dynamics can be seen. We believe that the trade-off between concepts and dynamics will be improved as the quality and size of text-video datasets increases in the future.

\subsection{Text-Image conditional video generation}
\begin{figure}[htb]
  \centering
  \includegraphics[width=1\linewidth]{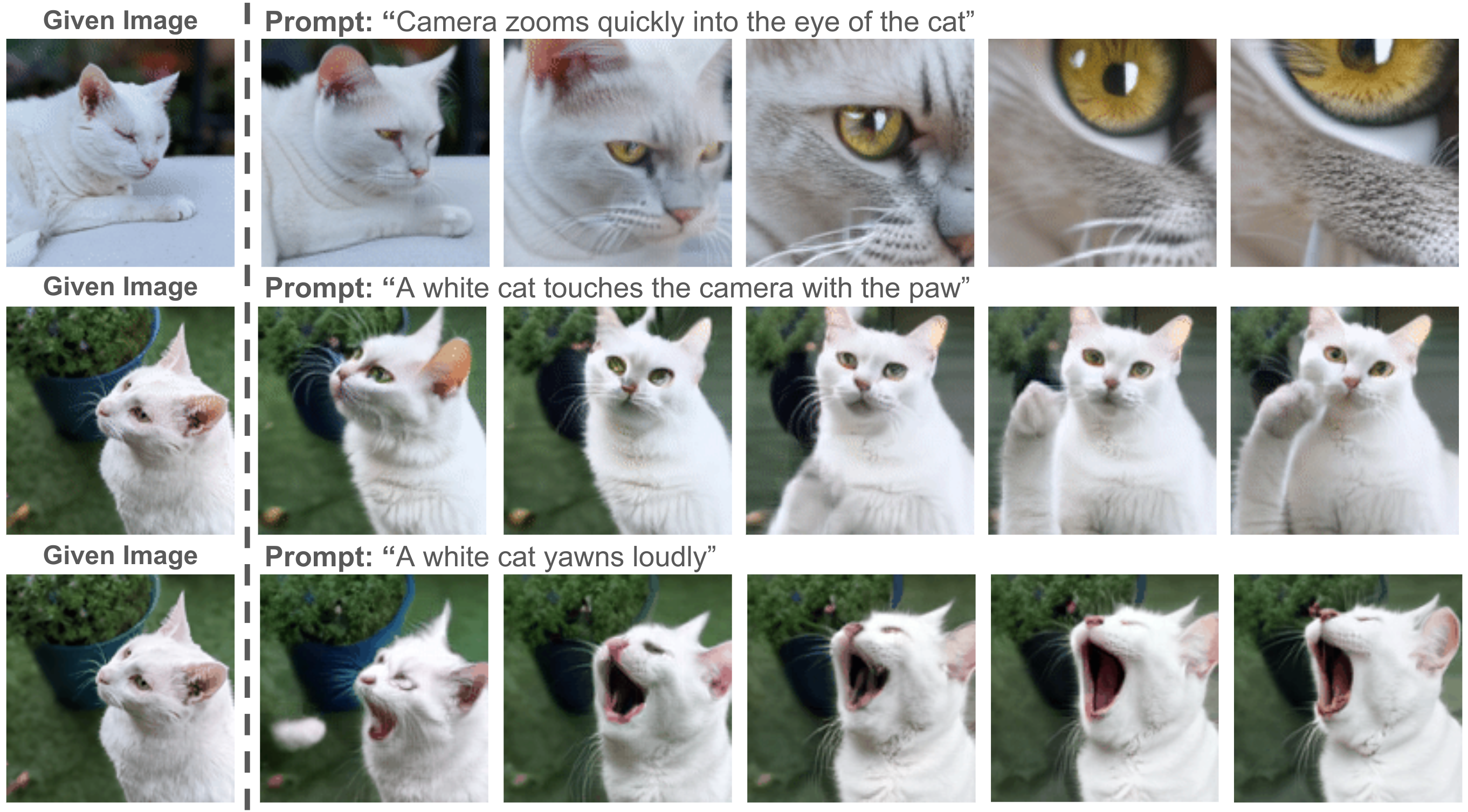}
  \caption{Animating images conditioned on a prompt. Each row demonstrates multiple frames of a generated video conditioned on a given first frame as well as a given text prompt. The first frames are new (captured by author's phone) and not observed during the training. The model \textit{animates} the given image while following the prompt. Full videos are available at \website.}
  \label{fig:cat}
\end{figure}

Given that \phenaki can be conditioned on both still images and text, an interesting setup is to \textit{animate} existing images given a text prompt. For this experiment, we use the same model from Section~\ref{sec:t2v} but conditioned on unseen pictures (captured with our phones from local subjects) and a related prompt. As it can be seen in Figure~\ref{fig:cat} the model can generate coherent videos starting from the given images, while following the given prompts. 

\subsection{Visual story telling by dynamic text inputs}
A notable and useful feature of \phenaki is that it is auto-regressive in time. This allows for generating long videos, while the prompt changes over time. Time variable prompts can be thought of as a \textit{\story}; a narration of the entire video where each prompt corresponds to a scene from the video. This allows for creating dynamically changing scenes. To the best our knowledge, this paper is the first work to generate such videos. 
An example of this can be seen in Fig.~\ref{fig:story} and on \website. The way it works is that we generate a video with the first prompt and then extend it in time by conditioning a possibly new prompt and on the last $N$, typically 5, previously generated frames.

\subsection{Video Encoding}
\label{sec:vid_quant}
To evaluate the video encoding and reconstruction performance of \cvivit, we use the Moments-in-Time (MiT)~\citep{monfortmoments} dataset. MiT contains ${\sim}802$K training, ${\sim}33$K validation and ${\sim}67$K test videos at 25 FPS. The MiT dataset, in contrast to other publicly available video datasets, is a high quality balanced dataset with high coverage and density of verbs depicting moments of a few seconds~\cite{monfortmoments}. We compare \cvivit against per-frame image based encoder-decoders that have been used as video quantizers for conditional video generation~\citep{harp,nuwa,cogvideo,nuwa,cogvideo,godiva}: a ViT \cite{vit_vqgan} and a convolutional VQ-GAN\cite{vqgan}. The experimental details can be found in the Appendix~\ref{app:sec:vid_quant}. 

\begin{table}
\small
\caption{\small Video reconstruction results on Moments-in-Time. The number of tokens is computed for 10 frames with the exception of \cvivit which is for 11, due to the isolated initial frame.}
\vspace{-.3cm}
\label{tab:quant}
\begin{center}
\begin{tabular}{lccc}
Method  & FID $\downarrow$ & FVD $\downarrow$ & \makecell{Number of Tokens} $\downarrow$\\
\hline
Conv VQ-GAN \cite{vqgan}                 & 7.5 & 306.1 & 2560 \\
Conv VQ-GAN + Video loss            & 13.7 & 346.5 & 2560 \\
ViT VQ-GAN \cite{vit_vqgan}                  & 3.4 & 166.6 & 2560 \\
ViT VQ-GAN + Video loss             & 3.8 & 173.1 & 2560 \\ \hline
% ViViT VQ-GAN (Ours)                 & 4.3 & 23.76 & 1280 & N \\
C-ViViT VQ-GAN (Ours)               & 4.5 & 65.78 & 1536 \\
\end{tabular}
\end{center}
\end{table}

% \begin{table}
% \small
% \vspace{-0.7cm}
% \caption{\small Video reconstruction results on Moments-in-Time. The number of tokens is computed for 10 frames with the exception of \cvivit which is for 11, due to the isolated initial frame.}
% \vspace{-.3cm}
% \label{tab:quant}
% \begin{center}
% \begin{tabular}{lcccc}
% Method  & FID $\downarrow$ & FVD $\downarrow$ & \makecell{Number of Tokens} $\downarrow$ & Supports Variable Length Videos\\
% \hline
% Conv VQ-GAN \cite{vqgan}                 & 7.5 & 306.1 & 2560 & Y\\
% Conv VQ-GAN + Video loss            & 13.7 & 346.5 & 2560 & Y \\
% ViT VQ-GAN \cite{vit_vqgan}                  & 3.4 & 166.6 & 2560 & Y \\
% ViT VQ-GAN + Video loss             & 3.8 & 173.1 & 2560 & Y \\ \hline
% % ViViT VQ-GAN (Ours)                 & 4.3 & 23.76 & 1280 & N \\
% C-ViViT VQ-GAN (Ours)               & 4.5 & 65.78 & 1536 & Y \\
% \end{tabular}
% \end{center}
% \end{table}

As demonstrated in Table~\ref{tab:quant}, we evaluate the video reconstruction quality using FID~\cite{heusel2017gans} and FVD~\cite{unterthiner2018towards}. Both FID and FVD compare the distribution of generated videos (or images) to the ground truth distribution. The FID ignores temporal coherency, while the FVD measures how well the spatio-temporal dynamics of the videos are reconstructed. Results in Table~\ref{tab:quant} show that per-frame image based methods slightly outperform our video method (indicated by marginally higher FID of \cvivit), however, they do poorly at modeling the spatio-temporal dynamics in video (significantly lower FVD of \cvivit). This is expected as \cvivit has spatio-temporal connections between patches in each frame, allowing space and time to be modeled together. In addition, \cvivit compresses the video into fewer tokens per video compared to the image based baselines. This is crucial as the number of tokens drastically impacts the computational cost of the transformer in downstream tasks. Furthermore, \cvivit tokens are auto-regressive in time which enables variable length videos to be modeled with the same encoder which is important for video extrapolation conditioned on previously generated frames.
% So, while \vivit achieves lower FID and FVD, \cvivit is more useful in practice.

\subsection{Image conditional video generation a.k.a Video prediction}
\label{sec:vid_bair}

\begin{table}
    \small
    \vspace{-.5cm}
    \begin{minipage}{.55\linewidth}
        \caption{\small Video prediction on Kinetics-600~\cite{Kinetics600}. While \phenaki is not designed for video prediction it achieves comparable results with SOTA video prediction models.}
        \vspace{-.3cm}
        \begin{center}
            \begin{tabular}{lr}
            \label{tab:kinetics}
                Method  & FVD $\downarrow$ \\
                \hline
                Video Transformer \cite{videotranf}            & 170.0 $\pm$ 5.00 \\
                CogVideo \cite{cogvideo}                    & 109.2 \quad \quad \ \ \ \ \\
                DVD-GAN-FP \cite{clark2019adversarial}                  & 69.1 $\pm$ 0.78 \\
                Video VQ-VAE \cite{videovqvae}            & 64.3 $\pm$ 2.04 \\
                CCVS  \cite{ccvs}                         & 55.0 $\pm$ 1.00 \\
                TrIVD-GAN-FP \cite{trIVD}                 & 25.7 $\pm$ 0.66 \\
                Transframer  \cite{transframer}           & 25.4  \quad \quad \ \ \ \ \\
                RaMViD \cite{diff4pred}                   & 16.5  \quad \quad \ \ \ \ \\
                Video Diffusion \cite{ho2022video}          & 16.2 $\pm$ 0.34 \\
                \hline
                Phenaki (Ours)                            & 36.4 $\pm$ 0.19 \\
            \end{tabular}
        \end{center}    
    \end{minipage}\hspace{0.2cm}%
    \begin{minipage}{.45\linewidth}
        \caption{\small Video prediction on BAIR~\cite{BAIRRobotPush}.}
        \vspace{-.3cm}
        \begin{center}
            \begin{tabular}{lr}
            \label{tab:bair}
                Method  & FVD $\downarrow$ \\
                \hline
                DVD-GAN \cite{clark2019adversarial}                & 109.8 \\
                VideoGPT \cite{videogpt}             & 103.3 \\
                TrIVD-GAN \cite{trIVD}               & 103.3 \\
                Transframer  \cite{transframer}      & 100.0 \\
                HARP \cite{harp}                     & 99.3 \\
                CCVS  \cite{ccvs}                    & 99.0 \\
                Video Transformer \cite{videotranf}  & 94.0 \\
                FitVid \cite{fitvid}                 & 93.6 \\
                MCVD \cite{mcvd}                     & 89.5 \\
                NUWA \cite{nuwa}                     & 86.9 \\
                RaMViD  \cite{diff4pred}             & 84.2 \\
                \hline
                Phenaki (Ours)                       & 97.0 \\
                % Phenaki (Ours) - 100 samples & 70.9 \\
                % Video Diffusion \cite{videodiff} - 100 samples & 66.9\\
            \end{tabular}
        \end{center}

    \end{minipage}
    
\end{table}
To evaluate the learnt video representation of \cvivit beyond reconstruction, we test it on the task of frame-conditioned video generation, also commonly known as video prediction~\cite{fitvid}. In this experiment, we test \phenaki on BAIR Robot Pushing benchmark~\cite{BAIRRobotPush} where the task is to generate 15 frames conditioned on a given single frame. For open domain videos, we test \phenaki on Kinetics-600~\cite{Kinetics600} where the task is to predict 11 frames given 5 frames. More details about these experiments can be found in Appendix~\ref{app:sec:vid_bair}. \cref{tab:bair,tab:kinetics} show the results of these experiments. Note that \phenaki is not specifically designed for video prediction, therefore, it lacks components such as skip connections in U-Nets which are known to improve the performance for video prediction methods~\cite{svg,villegas2019high,fitvid}. Nevertheless, our method is competitive on these benchmarks with SOTA video prediction methods. Overall, these experiments show that \phenaki is strong at modeling dynamics of the videos which is required for generating coherent videos from text.
%, by generating multiple frames in the future purely in the spatio-temporal latent space learned by our \cvivit encoder.

% \begin{itemize}
%     \item We show that our method is competitive at video prediction with SOTA
%     \item Our method is NOT made with video prediction specific design choices such as skip connections which help video prediction.
%     \item Our method is mainly designed with representation learning in mind for text-to-video synthesis
% \end{itemize}

% \begin{itemize}
%     \item Show we are better than NUWA at temporal dynamics due to cvivit as shown by FID-vid in zero-shot setting (we may want to have a model ready for the same datasets during rebuttal.
%     \item Show interesting things from` our model: Effects of CFG, short text-to-video examples, text-and-image-to-video
%     \item Depending on our results from Pieter's models, show study of using images and videos during training
%     \item Depending on our results from Pieter's models, evaluate on FID for text2image (on top of FVD on text2-video)
% \end{itemize}

% -- As we increase CFG scale \\
% - Training with video + images improves text-video alignment (CLIP and VideoCLIP score) \\
% - Training with t5x embeddings + images + video improves text-video alignment (CLIP and VideoCLIP) over training without t5x embeddings \\
% - Training and evaluating on the same dataset (LMS Video) seems to be the best on FVD \\

% -- We should compute first frame FID only to disambiguate the FVD results above when training and testing on lms video

\section{Related Works}

This paper is closely related to auto-regressive methods for text conditioned image and video generation. DALL-E~\cite{DALLE} \textit{translates} text tokens to discrete image embeddings learnt using a VQVAE~\cite{vqvae}. Parti~\cite{PARTI} has a similar architecture but can generate higher quality images by predicting tokens from a ViT-VQGAN~\cite{vit_vqgan} using a 21B parameters transformer. Similar architectures have been used for generating videos as well. GODIVA~\cite{godiva} uses a transformer to map text tokens to video tokens from a image based VQVAE. Given the large number of tokens from multiple frames, GODIVA relied on a local-attention mechanism. Similarly, NUWA~\cite{nuwa} and NUWA-Infinity~\cite{wu2022nuwa} both employ auto-regressive architectures to generate videos and images from text. NUWA generates fixed size outputs, while NUWA-Infinity introduces a second layer of auto-regressive computation to support variable size videos. Likewise, CogVideo~\cite{cogvideo} argues the main reason behind low quality video generation is the scarcity of good text-video data and tried to leverage pre-trained text to images models to generate high quality video. 

While \model sticks to the same architecture principles, it has major differences with previous work. Most notably, NUWA, NUWA-Infinity and CogVideo treat videos as a sequence of independent images. This can lead to poor modeling of dynamics and generate motion artifacts. To combat this, NUWA-infinity used the previous frame during decoding to combat this. In \model, we go further and treat videos as a temporal sequence of images which substantially decreases the number of video tokens given the redundancy in video generation, and results in a much lower training cost. The auto-regressive nature of the \phenaki also allows us to effectively condition on previous frames and generates longer videos as detailed in Section~\ref{sec:method}.

%\mbz{I'm not sure how much detail we wanna put here. commenting out for now.}
% Second, \model does not rely on an auto-regressive transformer. 
% \begin{itemize}
%     \item It is similar to maskgit
%     \item It is much faster in inference
%     \item Inference cost can actually be tuned, we can trade off quality for generation speed
%     \item The efficient maskgit inference opens the door to story mode. Which would be harder with regular transformers. 
% \end{itemize}

% Finally, 
% \begin{itemize}
%     \item CogVideo does pre-training
%     \item Nuwa does train on text to image and text2video jointly + a host of other tasks. In our model, text2video and text2image are actual naturally co-existing leading to a less complicated setup.
%     \item similar to imagen and video diffusion we use a pre-trained t5x text encoder. This difffers
% \end{itemize}

%Second, \model is the first model that utilizes joint training on images and videos at the same time to rectify issues with text-video datasets. Finally, \model is only about 1B parameters in size, which is much smaller than CogVideo at 9B and NUWA which only has a 800M parameters decoder. This smaller size, clearly highlights the importance of main contributions of our paper.
% \pikinder{Nuwa Looks to be 870M in total }

Diffusion models are another class of models which recently have been used for conditional and unconditional video generation, which we call VDM~\cite{ho2022video}. In VDM, authors proposed replacing the conventional U-Net architectures for 2D image modeling with a 3D space-time model to run the diffusion process directly on pixels. % VDM then uses two models to generate videos: one to generate every 4 frames and one to fills in between. 
While this approach provides an effective formulation for modeling videos, it is limited to fixed size videos. To address this issue, VDM provides an auto-regressive extension, which allows the model to generate longer videos but it is typically impractical due to high sampling time of diffusion models.

Text conditional video generation is a relatively new field of research, nonetheless, image conditional video generation, commonly known as video prediction, and unconditional video generation have been studied more comprehensively. These papers include deterministic methods using a combination of recurrent and convolutional networks~\cite{ranzato2014video,srivastava2015unsupervised,finn2016unsupervised, wang2017predrnn}, variational based stochastic methods~\cite{babaeizadeh2017stochastic,svg,villegas2019high,fitvid} and more recently by learning a discrete representation~\cite{videovqvae,rakhimov2020latent,transframer}, auto-regressive models~\cite{videotranf,videogpt,ccvs,harp}, diffusion models~\cite{mcvd,harvey2022flexible,yang2022diffusion,diff4pred} flow based models~\cite{kumar2019videoflow}, and finally adversarial based methods~\cite{vondrick2016generating,saito2017temporal,tulyakov2018mocogan,clark2019adversarial,saito2020train,trIVD}. These works mostly consider limited domain (e.g. robotic videos) prediction/generation, or short fixed size clips. Section~\ref{sec:exp} provides comparison with some of these models.

\section{Conclusion}

We introduced \phenaki, a model which is capable of generating variable length videos conditioned on a sequence of open domain text prompts. \phenaki uses \cvivit as video encoder. \cvivit is a new model which provides temporal-spatial compression while being auto-regressive in time.
The \cvivit model is a crucial part of \phenaki that allows it to generate variable length videos. We demonstrate how joint training on images and videos can improve the generation quality, and diversity, given the existence of much larger image-text dataset with order of magnitude more samples. The \phenaki model achieves good performance on video prediction, it can be used as to generate long videos conditioned on a text prompt. Additionally it is able to condition on both text and a starting frame. Finally, \phenaki is not limited to generating a video depicting a single concept or caption. It is actually able to generate longer coherent video stories based on a sequence of text prompts. The more complex narratives it can visualize demonstrate how this can become a great creative tool for story telling.

\section*{Ethics Statement}
While we have not explored potential downstream applications of the generative models described in this work, we believe Phenaki can have a positive impact in a variety of creative settings. In general, many of the samples from the model will not perfectly correspond to the input caption or the user’s intent; however, the end-user is likely to gain considerable time savings even if only one of the generated samples aligns with their intent. We thus foresee Phenaki being useful in eventually empowering users to accelerate their creativity, especially since the model can so quickly generate videos. Phenaki and similar models will be part of an ever-broad toolset for artists and non-artists alike, providing new and exciting ways to express creativity.

The flip-side of this acceleration and ease-of-use is the potential for harmful impact, as with many of the prior or concurrent work in generative modeling. An easy-to-use system like Phenaki can be repurposed for generating maliciously fake content and enable spreading of such content much easier. While the quality of the videos generated by Phenaki is not yet indistinguishable from real videos, getting to that bar for a specific set of samples is within the realm of possibility, even today. This can be particularly harmful if Phenaki is to be used to generate videos of someone without their consent and knowledge.

Like DALLE-2~\cite{ramesh2022hierarchical}, Imagen~\cite{saharia2022photorealistic}, Parti~\cite{PARTI} and others, Phenaki is trained on a collection of datasets that is known to encode a number of undesirable biases. LAION-400M~\cite{schuhmann2021laion} specifically has a variety of issues regarding violence, pornography, gore. While our primary image and video datasets have minimal traits like this, we did incorporate LAION-400M into our training and observed better results. In a currently training version of Phenaki, we use a set of datasets that minimizes such problems.

Taken together, these issues contribute to our decision not to release the underlying models, code, data or interactive demo at this time. Before we can do that, we want to focus our efforts on better understanding of data, prompt and output filtering. We would also like to more explicitly measure the biases encoded in the outputs of Phenaki, so that we can further mitigate them actively, either in the data, models or pre/post-processing steps.

% \subsubsection*{Author Contributions}
% Everyone was awesome!

\section*{Acknowledgments}
We would like to thank Niki Parmar for initial discussions. Special thanks to Gabriel Bender and Thang Luong for reviewing the paper and providing constructive feedback. We appreciate the efforts of Kevin Murphy and David Fleet for advising the project and providing feedback throughout. We are grateful to Evan Rapoport, Douglas Eck and Zoubin Ghahramani for supporting this work in a variety of ways. Tim Salimans and Chitwan Saharia helped us with brainstorming and coming up with shared benchmarks. Jason Baldridge was instrumental for bouncing ideas. Alex Rizkowsky was very helpful in keeping things organized, while Erica Moreira and Victor Gomes ensured smooth resourcing for the project. Sarah Laszlo and Kathy Meier-Hellstern have greatly helped us incorporate important responsible AI practices into this project, which we are immensely grateful for. Finally, Blake Hechtman and Anselm Levskaya were generous in helping us debug a number of JAX issues.

\bibliography{_references}
\bibliographystyle{plainnat}

\newpage
\appendix
\section{Hyper-Parameters}
\begin{table}[hbtp]
    \centering
    \begin{tabular}{l|c|l}
         Symbol & Value & Description \\\hline
         $t_x, w_x, h_x, c_x$ & $11, 128, 128, 3$ & Video dimensions \\
         $t_p, w_p, h_p, c_p$ & $2, 8, 8, 3$ & Patches dimensions (all frames except the first one) \\
         $t_z, w_z, h_z$ & $6, 16, 16$ & Video tokens dimension (before linear projection) \\
         $h_z$ & 512 & Hidden size in the transformer layer \\
         $d_z$ & 32 & Embedding dimension (after linear projection) \\
         $-$ & 4 & Number of layers for spatial transformer \\
         $-$ & 4 & Number of layers for temporal transformer \\
         $-$ & 2048 & MLP size \\
         $|E|$ & 8192 & Codebook size \\ 
         \hline
         - & AdamW &  Optimizer \\
         $\beta_1$ & 0.9 & first moment of gradient \\
         $\beta_2$ & 0.99 & second moment of gradient \\
         - & 1e-4 & Learning rate \\ 
         - & 1e-4 & Weight decay \\ 
         - & Cosine decay & Learning rate scheduler \\ 
         - & 1M & Target number of training steps for learning rate scheduler \\
         - & 100K & Warmup steps \\
         - & 10 & Gradient clipping magnitude \\
         - & 1028 & Batch size \\
         
    \end{tabular}
    \caption{Hyperparamters used for \cvivit architecture and optimizer.}
    \label{tab:hparams_cvivit}
\end{table}

\begin{table}[hbtp]
    \centering
    \begin{tabular}{l|c|l}
         Symbol & Value & Description \\\hline
         $|\bz|$ & 1536 & Sequence Length \\
         - & 24 & Number of layer \\
         - & 2048 & Embedding dimension \\
         - & 8192 & MLP dimension \\
         - & 32 & Number of heads \\
         \hline
         - & AdamW &  Optimizer \\
         $\beta_1$ & 0.9 & first moment of gradient \\
         $\beta_2$ & 0.99 & second moment of gradient \\
         - & 1e-4 & Learning rate \\ 
         - & 1e-4 & Weight decay \\ 
         - & Cosine decay & Learning rate scheduler \\ 
         - & 4M & Target number of training steps for learning rate scheduler \\
         - & 10K & Warmup steps \\
         - & 10 & Gradient clipping magnitude \\
         - & 512 & Batch size \\
    \end{tabular}
    \caption{Hyperparamters used for MaskGIT architecture and optimizer.}
    \label{tab:hparams_maskgit}
\end{table}

\section{Details of Experiments}

\subsection{Video Quantization}
\label{app:sec:vid_quant}
\subsubsection{Network architecture}
All encoder-decoder baselines have approximately $50\text{M}$ parameters.
The Convolutional baseline encoder architecture consists of $5$ convolutional blocks with channel multipliers of $[1, 1, 2, 2, 4]$, $2$ residual layers and $128$ hidden units per block, and embedding dimension of 256.
The ViT baseline encoder architecture consists of an image patchification step over non-overlapping $8\times8$ spatial patches which are linearly transformed into image tokens.
Next, we follow with $8$ transformer layers with $512$ hidden units, $8$ attention heads, $2048$ mlp units, and embedding dimension of $32$.
\cvivit encoder architecture patches the first frame to non-overlapping $8\times8$ patches, and then the rest of the frames to non-overlapping $2\times8\times8$ spatio-temporal patches which are linearly transformed into video embeddings.
Next, \cvivit encoder architecture consists of $4$ spatial and $4$ temporal transformer layers with $512$ hidden units, $8$ attention heads, $2048$ mlp hidden units, and embedding dimension of $32$.
The decoder architecture for all models is the same as the encoder but in reverse to put the latent embeddings back to image space.
The VQ objective is trained with commitment loss of $\beta=0.25$ and codebook size of $8192$.
The discriminator architecture is the StyleGAN \cite{stylegan} discriminator with blur resample, and channel multiplier of $1$.

\subsubsection{Training}
We train all encoder-decoder baselines and with StyleGAN~\citep{stylegan} discriminators with a batch size of $128$ using Adam optimizer~\cite{adam} with $\beta_1=0.9$ and $\beta_2=0.99$. We use a linear learning rate warmup to a peak value of $1 \times 10^{-4}$ over $100,000$ steps and then decaying over the remaining $900,000$ steps with a cosine schedule, and use a decoupled weight decay \cite{adamw} of $1 \times 10^{-4}$ for the encoder-decoder and discriminator. To capture longer time horizons during training and better evaluate temporal coherence, we downsample the MiT dataset from 25 FPS to 6 FPS and evaluate on videos of 10 frames at spatial resolution of $128\times128$.

\subsection{Image conditional video generation}
\label{app:sec:vid_bair}

\subsubsection{BAIR Robot Push \cvivit architecture} \label{app:sec:bair}
We use a similar setup as in Section \ref{app:sec:vid_quant}, but the video tokenization step is done over $4\times4$ spatial patches on the first image and $2\times4\times4$ spatio-temporal patches in the rest of the video. The spatial encoder consists of 8 layers and the temporal encoder consists of 6 layers.

\subsubsection{Kinetics-600 \cvivit architecture}
We use a similar setup as in Section \ref{app:sec:bair}, but both the spatial encoder and temporal encoder consist of 8 layers.

\subsubsection{MaskGIT architecture}
To perform video prediction in latent space in the BAIR Robot Push and Kinetics-600 datasets, we use an unconditional transformer architecture consisting of $24$ layers, $768$ hidden
units, $16$ attention heads, dropout and attention dropout rate of $0.1$, $3072$ mlp hidden units.

\subsubsection{Training and Inference}
As described in Table~\ref{tab:hparams_maskgit}, we train \cvivit with the same optimizer setup as in Sec~\ref{app:sec:vid_quant}, but we do not downsample the FPS of any of the datasets in this section for fair comparison with the video prediction baselines. We train MaskGIT on the video tokens extracted using \cvivit in an unconditional setting, that is, we do not assume frames or text inputs to be given. During training, we use the Adam \cite{adam} optimizer with $\beta_1=0.9$ and $\beta_2=0.99$. We use a linear learning rate warmup up to a peak value of $1\times10^{-4}$ over $10,000$ steps, and constant learning rate schedule for ${~\sim}2M$ steps. At inference time, we initialize MaskGIT given a number of input frames, and predict the rest of the frames depending on the dataset on which we evaluate.

\subsection{Text conditional video generation}
\label{app:sec:text2vid}

\subsubsection{Architecture}
In our text conditional video generation, we use the same \cvivit architecture and training described in Section \ref{app:sec:vid_quant}.
To train MaskGIT, we include a text conditioning in the form of T5X embeddings \cite{roberts2022scaling} which are used as input through the use of cross attention with the video tokens. 
We reduce the number of parameters of our base model for fairness in the quantitative comparisons against NUWA. The MaskGIT architecture used against NUWA consists of $20$ transformer layers with $1536$ hidden units, $24$ attention heads, and $6144$ MLP hidden units, resulting in $0.9\text{B}$ parameters similar to NUWA. For the main experiments in this paper, we use a larger architecture that consists of consists of $24$ transformer layers with $2048$ hidden units, $32$ attention heads, and $8192$ mlp hidden units, resulting in $1.8\text{B}$ parameters.

\subsubsection{Training and inference}
For all our text-conditional video generation, we use the training parameters Table~\ref{tab:hparams_maskgit}.

\subsubsection{Inference parameters against NUWA}
We use $\lambda=0.1$, $12$ MaskGIT iterations, and temperature of $4.0$. 

\subsubsection{Inference parameters for ablation of image and video data for training.}
We use $\lambda=6$, $24$ MaskGIT iterations, and temperature of $4.0$. 

\subsubsection{Inference parameters for all videos in the paper.}
We use $\lambda=12$, $48$ MaskGIT iterations, and temperature of $8.0$. 

\begin{figure}[htb]
  \centering
  \includegraphics[width=0.9\linewidth]{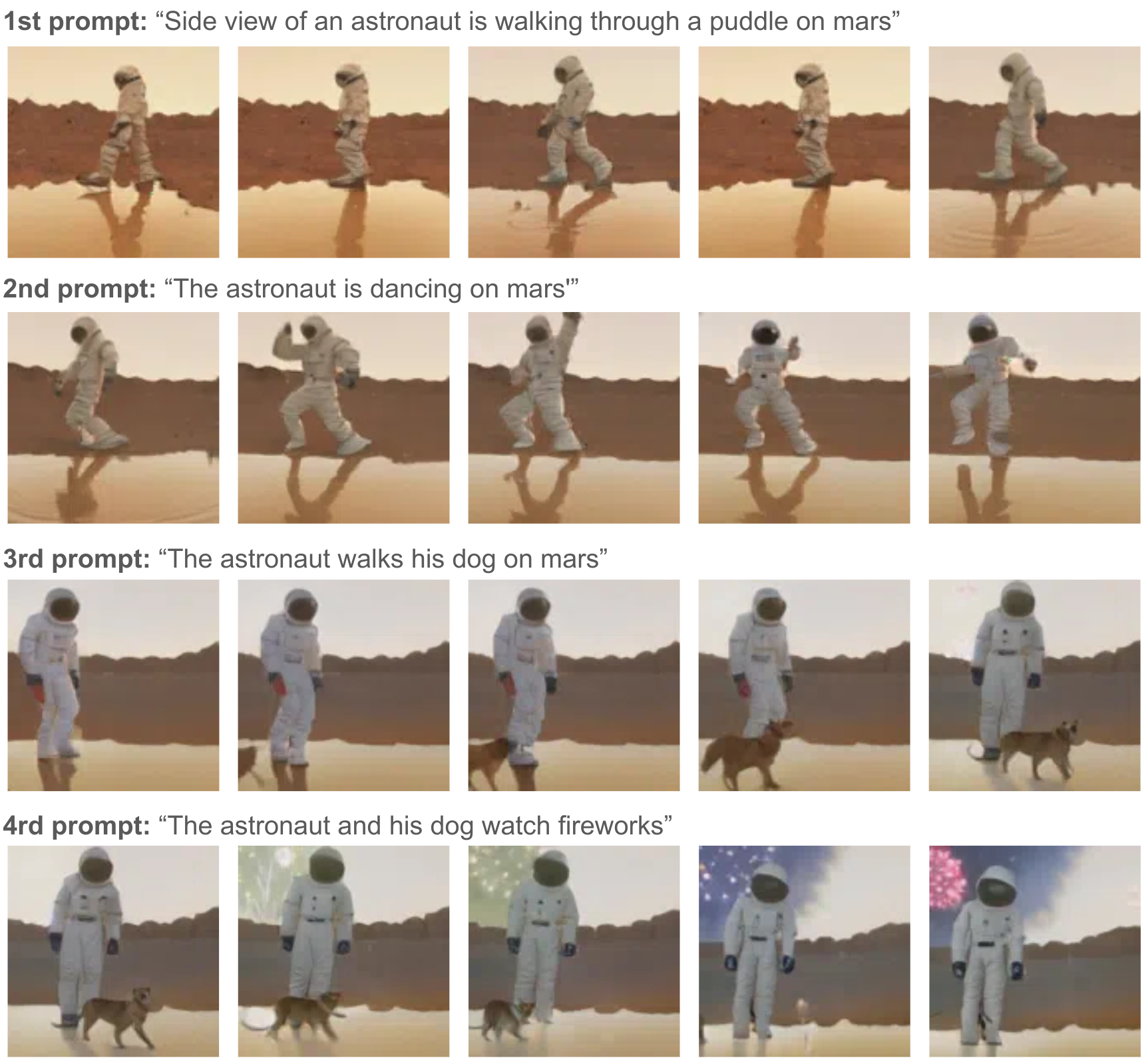}
  \caption{Another example of \story conditional video generation. Full videos are available at \website.}
  \label{fig:story2}
\end{figure}

\end{document}